\documentclass{article}
\usepackage{spconf,amsmath,graphicx}
\usepackage{multirow}
\usepackage{cite}
\usepackage{booktabs}
\usepackage{amssymb}
\usepackage{array}

\usepackage{xcolor}
\usepackage{hyperref}

\title{DistilHuBERT: Speech Representation Learning\\by Layer-wise Distillation of Hidden-unit BERT}
\name{Heng-Jui Chang, Shu-wen Yang, Hung-yi Lee \thanks{We thank to National Center for High-performance Computing (NCHC) of National Applied Research Laboratories (NARLabs) in Taiwan for providing computational and storage resources.
}}
\address{College of Electrical Engineering and Computer Science, National Taiwan University}

\begin{document}
\ninept
\maketitle
\begin{abstract}
Self-supervised speech representation learning methods like wav2vec 2.0 and Hidden-unit BERT (HuBERT) leverage unlabeled speech data for pre-training and offer good representations for numerous speech processing tasks.
Despite the success of these methods, they require large memory and high pre-training costs, making them inaccessible for researchers in academia and small companies.
Therefore, this paper introduces DistilHuBERT, a novel multi-task learning framework to distill hidden representations from a HuBERT model directly.
This method reduces HuBERT's size by 75\% and 73\% faster while retaining most performance in ten different tasks.
Moreover, DistilHuBERT required little training time and data, opening the possibilities of pre-training personal and on-device SSL models for speech.
\end{abstract}
\begin{keywords}
Self-supervised learning, speech representation learning, knowledge distillation, model compression
\end{keywords}

\section{Introduction}
\label{sec:intro}

Recently, self-supervised learning (SSL) methods for speech representation learning succeeded in many tasks \cite{liu2020mockingjay,liu2021tera,ling2020decoar,ling2020decoar2,liu2021npc,chung2019apc,chung2020vq-apc,chung2020improvedapc,oord2018cpc,riviere2020m-cpc,schneider2019wav2vec,baevski2020vq-wav2vec,baevski2020wav2vec2,sadhu2021wav2vec-c,hsu2021hubert,pascual2019pase,ravanelli2020paseplus,baevski2019effectiveness}.
They learn from targets derived from unlabeled speech data.
They can be roughly categorized into two classes by learning objectives: generative and discriminative.
For generative methods, models try to either reconstruct masked acoustic features \cite{liu2020mockingjay,liu2021tera,ling2020decoar,ling2020decoar2,liu2021npc}, or generate future acoustic features \cite{chung2019apc,chung2020vq-apc,chung2020improvedapc}.
For the discriminative methods, models either learn by contrastive learning \cite{oord2018cpc,riviere2020m-cpc,schneider2019wav2vec,baevski2020vq-wav2vec,baevski2020wav2vec2,sadhu2021wav2vec-c} or classifying pseudo labels \cite{hsu2021hubert}.

Many previous SSL methods for speech representation learning are designed for few tasks like speech recognition.
Therefore, Speech processing Universal PERformance Benchmark (SUPERB) \cite{yang2021superb} and LeBenchmark \cite{evain2021lebenchmark} are developed to evaluate the effectiveness of these methods on general speech processing applications.
These benchmarks comprehensively assess the capability of pre-trained models by applying them to multiple downstream tasks like recognition, detection, semantics, and speaker identification.

Although recent developments of speech SSL methods show excellent performance across various tasks, many require large memory and high pre-training costs, including wav2vec 2.0~\cite{baevski2020wav2vec2} and HuBERT~\cite{hsu2021hubert}.
The limitations make these models unsuitable for on-device computation and make researchers in academia and small corporations difficult to use~\cite{hannun2021history}.
Knowledge distillation is a common method for compressing models \cite{hinton2015distilling}, in which a small student model is learned to generate the teacher model's outputs or hidden representations.
Distilling knowledge has shown to be effective for NLP, and DistilBERT \cite{sanh2019distilbert} and TinyBERT \cite{jiao2020tinybert} are good examples.
However, we found these approaches ineffective in distilling speech SSL models, and few studies investigated this problem \cite{peng2021shrinking,lai2021parp}.
Therefore, we propose a novel multi-task learning framework to layer-wise distill hidden representations of SSL speech models.

In this work, we distill HuBERT and obtain DistilHuBERT.
DistilHuBERT uses three prediction heads to respectively predict the 4$^{\text{th}}$, 8$^{\text{th}}$, and 12$^{\text{th}}$ HuBERT hidden layers' output.
After training, the heads are removed because the multi-task learning paradigm forces the DistilHuBERT model to learn representations containing rich information.
DistilHuBERT reduces HuBERT's size by 75\% and speedup by 73\%, retaining most performance and requiring less training time.
Moreover, we offer comprehensive analyses of proposed methods and show DistilHuBERT's ability to distill knowledge with few data.
To our knowledge, this is the first attempt to distill speech representations from SSL pre-trained models directly.

\section{Methods}
\label{sec:method}

\subsection{HuBERT}
\label{subsec:hubert}

The proposed distillation approach can be applied to any self-supervised model.
This paper takes HuBERT~\cite{hsu2021hubert} as the teacher because it outperforms other methods in almost all tasks in SUPERB~\cite{yang2021superb}, showing it offers good speech representations.
HuBERT consists of a CNN and a transformer encoder \cite{vaswani2017attention} to classify randomly masked frames to pseudo labels.
The labels are obtained by clustering MFCC or another model's hidden units.
Despite HuBERT's success, two major disadvantages appear.
First, HuBERT models have 95M to 1B parameters, consuming large memory and slow inference.
Second, the training costs are high, e.g., HuBERT Base pre-training requires 32 GPUs and 2k GPU hours.
These issues make academia and small corporations unable to pre-train their models or apply these technologies to products.
Hence, this paper aims to alleviate these problems.

\subsection{DistilHuBERT}
\label{subsec:framework}

\begin{figure*}[t]
    \centering
    \includegraphics[width=0.85\linewidth]{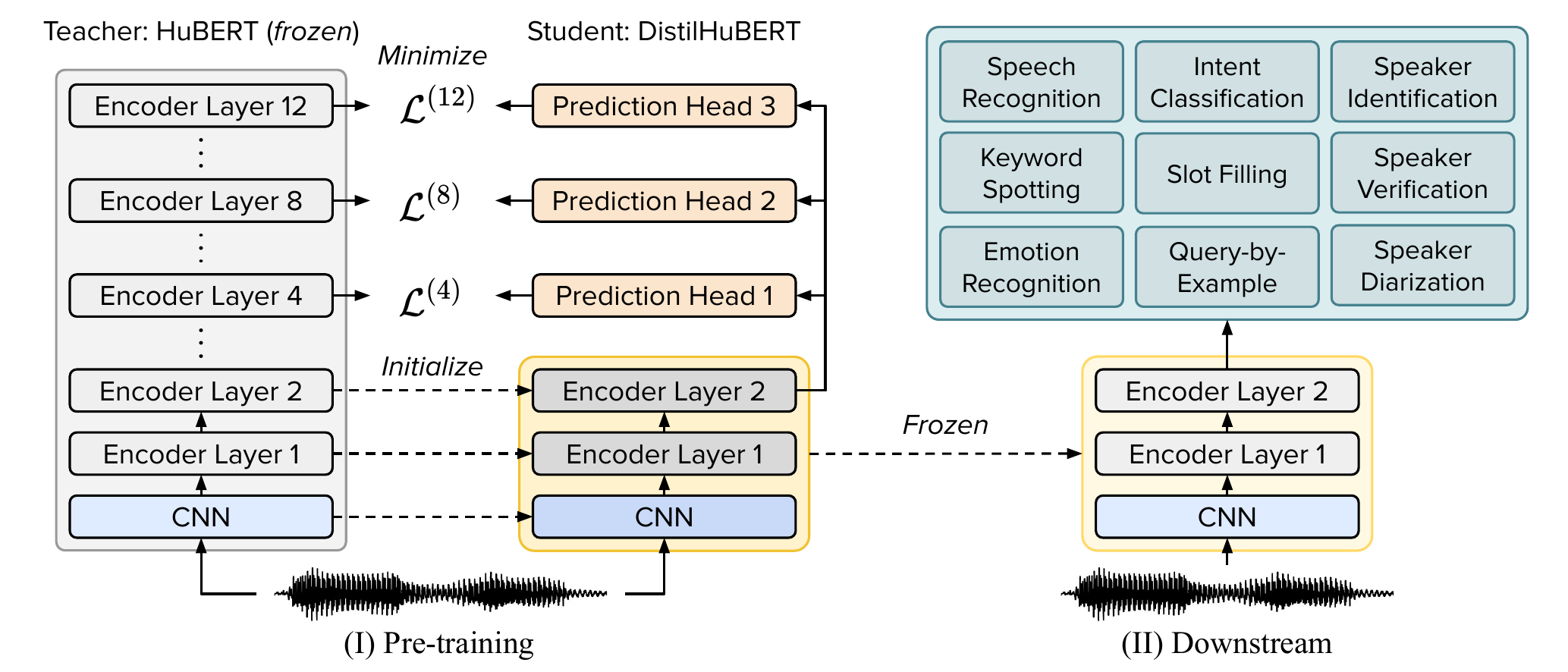}
    \vspace*{-10pt}
    \caption{
        The proposed DistilHuBERT framework.
        (I) Before pre-training, the student is initialized with the teacher's parameters.
        Then, student's prediction heads learn to generate the teacher's hidden representations by minimizing the loss function $\mathcal{L} = \mathcal{L}^{(4)} + \mathcal{L}^{(8)} + \mathcal{L}^{(12)}$.
        (II) The pre-trained DistilHuBERT model is frozen to generate speech representations for downstream tasks.
    }
    \label{fig:framework}
    \vspace*{-6pt}
\end{figure*}

In a typical knowledge distillation method, a student learns to generate the teacher's output.
However, different layers in self-supervised speech models contain different information like speaker identity or semantics \cite{baevski2020wav2vec2,pasad2021layer,chang2021exploration}.
An SSL model's output layer does not always offer rich information.
For instance, wav2vec 2.0 stores phonetic information in its middle layers \cite{pasad2021layer,baevski2021unsupervised}.
Therefore, the student only learning the last layer of the teacher might benefit little.
One possible way to solve this problem is to let different hidden layers of the student learn from different layers of the teacher \cite{jiao2020tinybert}.
The student model must be deep enough to have multiple layers, but against having a small model.

This paper proposes a novel teacher-student framework for speech representation learning by multi-task knowledge distillation, as shown in Fig. \ref{fig:framework}.
We distill HuBERT to obtain DistilHuBERT, consisting of a CNN feature extractor and a small transformer encoder.
The basic idea of our approach for knowledge distillation is to learn to generate multiple teacher's hidden representations from shared representations.
Therefore, we propose predicting teacher's hidden representations with separate prediction heads as shown in Fig.~\ref{fig:framework}(I).
This objective is a multi-task learning paradigm.
It encourages the transformer encoder to produce compact representations for multiple prediction heads.
After pre-training, we remove the heads.
The model parameters are frozen, and the output representations are used for various downstream tasks, as shown in Fig.~\ref{fig:framework}(II).

\noindent\textbf{Objective Function}.
Here, we describe the objective function of DistilHuBERT.
We denote the $D$ dimensional feature vectors at time $t^{\text{th}}$ produced by the teacher's $l^{\text{th}}$ layer and the student's corresponding prediction head respectively as $\hat{\boldsymbol{h}}_t^{(l)}$ and $\boldsymbol{h}_t^{(l)}$.
The loss function is
\begin{equation}
    \begin{split}
        \mathcal{L}^{(l)} & = \mathcal{L}_{\ell 1}^{(l)} + \lambda \mathcal{L}_{\cos}^{(l)} \\
        & = \sum_{t=1}^{T} \left[\frac{1}{D}\left\| \boldsymbol{h}^{(l)}_t - \hat{\boldsymbol{h}}^{(l)}_t \right\|_1 - \lambda \log\sigma\left( \cos\left(\boldsymbol{h}^{(l)}_t, \hat{\boldsymbol{h}}^{(l)}_t\right) \right) \right],
    \end{split}
    \label{eq:loss}
\end{equation}
where $T$ is the number of time steps.
$\sigma$ and $\cos (\cdot, \cdot)$ respectively denote sigmoid activation and cosine similarity.
Minimizing $\mathcal{L}^{(l)}$ is equivalent to simultaneously minimizing the $\ell1$ distance and maximizing the cosine similarity between the hidden representations.
We empirically found that considering both $\mathcal{L}_{\ell 1}$ and $\mathcal{L}_{\cos}$ yielded better performance than considering only one of them.
$\lambda >$ 0 controls the contribution of the cosine similarity loss.

\noindent\textbf{Parameter Initialization}.
Following DistilBERT \cite{sanh2019distilbert}, we initialized DistilHuBERT with HuBERT's CNN extractor and the first two transformer layers.

\noindent\textbf{Reducing Computation for Distillation}.
We note that the number of prediction heads can be 1 to $L$, where $L$ is the number of hidden layers in the self-supervised speech model to be distilled.  
Because neighboring layers' representations might contain similar information, the student model only predicts some specific layers in the teacher model and skip other layers in our implementation.
For example, in Fig.~\ref{fig:framework}, only the 4$^{\text{th}}$, 8$^{\text{th}}$, and 12$^{\text{th}}$ HuBERT hidden layers' outputs are predicted. 
In the experiments, we will show that predicting these three layers is sufficient.

\begin{table*}[t]
    \caption{
        Results on SUPERB \cite{yang2021superb}.
        The metrics include accuracy (Acc\%) phoneme error rate (PER\%), word error rate (WER\%), maximum term weighted value (MTWV), F1 score (F1\%), concept error rate (CER\%), equal error rate (EER\%), and diarization error rate (DER\%).
    }
    \vspace{2pt}
    \label{tab:superb}
    \centering
    \begin{tabular}{l @{~~}c@{~~} @{~}c@{~~} @{~}c@{~~} @{~}c@{~~} @{~}c@{~~} @{~}c@{~~} @{~}c@{~~} @{~}c@{~~} @{~}c@{~~} @{~}c@{~~} @{~}c@{~~} | c}
        \toprule
        & \# param. & PR & KS & IC & SID & ER & ASR (+LM) & QbE & SF & ASV & SD & \\
        \cmidrule{2-12}
        Method &  Millions &  PER$\downarrow$ &  Acc$\uparrow$ &  Acc$\uparrow$ &  Acc$\uparrow$ &  Acc$\uparrow$ &  WER$\downarrow$ &  MTWV$\uparrow$ &  F1$\uparrow$ / CER$\downarrow$ &  EER$\downarrow$ &  DER$\downarrow$ & Rank$\downarrow$ \\
        \midrule
        \textit{(I) Baselines} \cite{yang2021superb} \\
        \midrule
        FBANK & 0 & 82.01 & 8.63 & 9.10 & 8.5E-4 & 35.39 & 23.18 / 15.21 & 0.0058 & 69.64 / 52.94 & 9.56 & 10.05 & 8.8 \\
        TERA & 21.33 & 49.17 & 89.48 & 58.42 & 57.57 & 56.27 & 18.17 / 12.16 & 0.0013 & 67.50 / 54.17 & 15.89 & 9.96 & 7.9 \\
        wav2vec & 32.54 & 31.58 & 95.59 & 84.92 & 56.56 & 59.79 & 15.86 / 11.00 & 0.0485 & 76.37 / 43.71 & 7.99 & 9.90 & 6.5 \\
        DeCoAR 2.0 & 85.12 & 14.93 & 94.48 & 90.80 & 74.42 & 62.47 & 13.02 / 9.07 & 0.0406 & 83.28 / 34.73 & 7.16 & 6.59 & 4.1 \\
        HuBERT (teacher) & 94.68 & 5.41 & 96.30 & 98.34 & 81.42 & 64.92 & 6.42 / 4.79 & 0.0736 & 88.53 / 25.20 & 5.11 & 5.88 & 1.0 \\
        \midrule
        \multicolumn{3}{l}{\textit{(II) Distillation Baselines}} \\
        \midrule
        predict last layer & 24.67 & 16.71 & 96.07 & 95.57 & 43.44 & 62.81 & 13.67 / 9.43 & 0.0423 & 81.52 / 37.31 & 7.29 & 6.45 & 4.6 \\
        predict w/ hidden & 23.49 & 17.43 & 95.46 & 94.91 & 54.90 & 61.49 & 14.95 / 10.27 & 0.0529 & 83.73 / 35.14 & 7.41 & 6.86 & 5.0 \\
        \midrule
        \textit{(III) Proposed} \\
        \midrule
        w/ prediction heads & 27.03 & 14.35 & 96.20 & 94.17 & 72.83 & 62.73 & 13.26 / 8.99 & 0.0451 & 83.14 / 35.53 & 6.85 & 5.97 & 3.3 \\
        DistilHuBERT & 23.49 & 16.27 & 95.98 & 94.99 & 73.54 & 63.02 & 13.37 / 9.21 & 0.0511 & 82.57 / 35.59 & 8.55 & 6.19 & 3.8 \\
        \bottomrule
    \end{tabular}
    \vspace*{-8pt}
\end{table*}

\section{Experiments}
\label{sec:exp}

\begin{figure*}[t]
    \begin{minipage}[b]{.33\linewidth}
        \centering
        \centerline{\includegraphics[width=\linewidth]{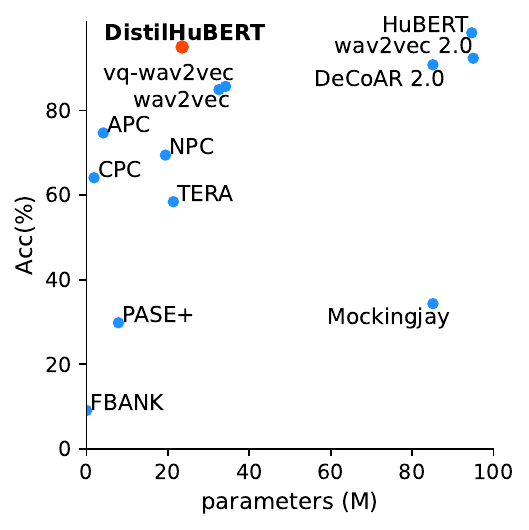}}
        \vspace*{-4pt}
        \centerline{(a) IC}\medskip
    \end{minipage}
    \hfill
    \begin{minipage}[b]{.33\linewidth}
        \centering
        \centerline{\includegraphics[width=\linewidth]{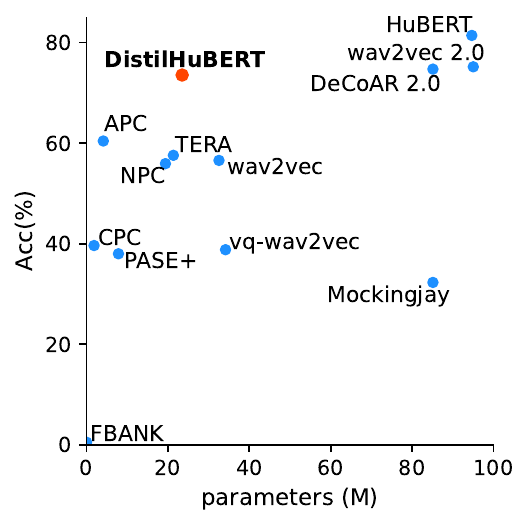}}
        \vspace*{-4pt}
        \centerline{(b) SID}\medskip
    \end{minipage}
    \hfill
    \begin{minipage}[b]{0.33\linewidth}
        \centering
        \centerline{\includegraphics[width=\linewidth]{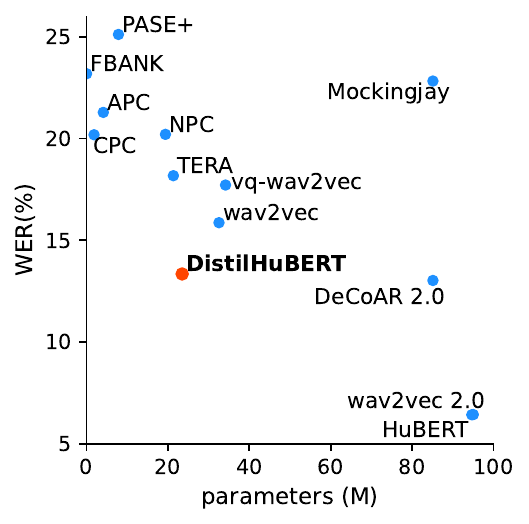}}
        \vspace*{-4pt}
        \centerline{(c) ASR (w/o LM)}\medskip
    \end{minipage}
    \vspace{-20pt}
    \caption{
        Comparisons of pre-trained model size vs. performance on three tasks: IC, SID, and ASR.
    }
    \label{fig:size_perform}
    \vspace{-10pt}
\end{figure*}

\subsection{Experimental Setup}
\label{subsec:setup}

Experiments were implemented with S3PRL \cite{liu2021tera,liu2020mockingjay} and fairseq \cite{ott2019fairseq}.

\noindent\textbf{Model}.
The HuBERT model had a 7-layer CNN and a 12-layer transformer encoder, while DistilHuBERT had a similar architecture, but its transformer encoder had only two layers.
DistilHuBERT had three prediction heads respectively predicting the 4$^{\text{th}}$, 8$^{\text{th}}$, and 12$^{\text{th}}$ HuBERT hidden layers.
We applied past distillation methods to DistilHuBERT for comparison \cite{sanh2019distilbert,jiao2020tinybert}.
As shown in Table \ref{tab:superb}, "predict last layer" predicted only the final output of HuBERT.
"Predict w/ hidden" was trained to predict the three chosen layers with its hidden layers, i.e., the CNN and the two transformer encoder layers respectively predicted the 4$^\text{th}$, 8$^\text{th}$, and 12$^\text{th}$ HuBERT layers.
"W/ prediction heads" preserved the three prediction heads after pre-training.

\noindent\textbf{Data}.
We used the 960-hour LibriSpeech dataset \cite{Panayotov2015libri} for knowledge distillation, except Sec.~\ref{subsec:diff_data} used the English Wall Street Journal (WSJ) \cite{Paul92wsj} and the Mandarin AISHELL-1 \cite{bu2017aishell} speech corpora.

\noindent\textbf{Pre-training}.
The default DistilHuBERT was trained on a 32GB V100 GPU for 200k updates with a batch size of 24 utterances, taking roughly 55 hours.
The learning rate was linearly increased to 2e-4 in the first 7\% of updates, then linearly decreased to zero.
$\lambda =$ 1 in Eq. (\ref{eq:loss}).
Note that training DistilHuBERT was cheaper than HuBERT since the HuBERT Base model required 2k GPU hours.

\subsection{SUPERB}
\label{subsec:superb}

DistilHuBERT was evaluated on SUPERB \cite{yang2021superb}.
Each SSL model was frozen, and a trainable weighted sum summarized the hidden representations.
Each downstream task used a simple model with minimal labeled data using the summarized representations.
The tasks included phoneme recognition (PR), keyword spotting (KS), intent classification (IC), speaker identification (SID), emotion recognition (ER), automatic speech recognition (ASR), query by example spoken term detection (QbE), slot filling (SF), automatic speaker verification (ASV), and speaker diarization (SD).

Results on SUPERB are shown in Table \ref{tab:superb}.\footnote{We only show some results because of the paper length limitation. A complete comparison is in \url{https://superbbenchmark.org/}.}
The rightmost column was each method's ranking score by averaging its ranks in the ten tasks, offering an easy way to compare the performance of the methods.
First, the ranking scores revealed that DistilHuBERT outperformed all methods, except HuBERT (its teacher), indicating DistilHuBERT retained most of HuBERT's performance (sec. (III) vs. (I)).
Next, the distillation baselines in section (II) performed worse than DistilHuBERT, especially in PR and SID tasks, showing the proposed framework distilled better representations using separate prediction heads.
Moreover, DistilHuBERT slightly degraded without the prediction heads, but it reduced the model size by 13\%, implying the prediction heads were redundant after pre-training.

Furthermore, Fig. \ref{fig:size_perform} visualizes the relation between the model sizes and their performance on IC, SID, and ASR.
We chose IC, SID, and ASR for comparison since they were easier to observe the performance differences.
In Fig. \ref{fig:size_perform}(a)(b), DistilHuBERT showed competitive performance compared with HuBERT and wav2vec 2.0 and surpassed all other similar-sized methods in IC and SID.
In Fig. \ref{fig:size_perform}(c), DistilHuBERT degraded more in ASR yet still better than most other SSL methods, showing the better trade-off between the model size and performance.

\begin{table}[t]
    \caption{
        Comparison of different SSL model sizes and inference time.
        The inference was performed on 4 CPUs extracting all features from the LibriSpeech dev-clean set with a batch size of one.
        Results were averaged over three runs.
    }
    \vspace{2pt}
    \label{tab:model_size}
    \centering
    \begin{tabular}{lcc}
        \toprule
        & \# param. & Inf. time \\
        \cmidrule{2-3}
        Model & Millions & seconds \\
        \midrule
        HuBERT & 94.68 (100\%) & 992 (1.00X) \\
        DeCoAR 2.0 & 85.12 (90\%) & 924 (1.07X) \\
        DistilHuBERT & 23.49 (25\%) & 574 (1.73X) \\
        \bottomrule
    \end{tabular}
    \vspace*{-8pt}
\end{table}

\subsection{Model Size and Inference Speed}

We compared the sizes and inference speed of several methods as shown in Table \ref{tab:model_size}.
DistilHuBERT had a significantly smaller model size than HuBERT and offered a 73\% speedup.
DistilHuBERT performed equally well as DeCoAR 2.0 but was smaller and faster.
Combining the results in Tables \ref{tab:superb} and \ref{tab:model_size}, DistilHuBERT was fast, small, and powerful.
Note that DistilHuBERT can be further compressed by pruning or quantization for on-device computation.

\subsection{Ablation Study}
\begin{table}[t]
    \caption{
        Ablation study of the techniques used.
    }
    \vspace{2pt}
    \label{tab:ablation}
    \centering
    \begin{tabular}{lccc}
        \toprule
        & IC & SID & ASR \\
        \cmidrule{2-4}
        Method & Acc$\uparrow$ & Acc$\uparrow$ & WER$\downarrow$ \\
        \midrule
        DistilHuBERT & 94.99 & 73.54 & 13.34 \\
        \midrule
        w/o $\mathcal{L}_{\cos}$ & 93.36 & 72.54 & 13.74 \\
        w/o teacher init & 94.20 & 73.68 & 13.41 \\
        \bottomrule
    \end{tabular}
    \vspace*{-10pt}
\end{table}

Table~\ref{tab:ablation} shows the effectiveness of the cosine similarity loss in Eq. (\ref{eq:loss}) and teacher initialization. 
Both techniques showed some benefits to training DistilHuBERT, and we thus used them in the rest of the experiments.
Removing cosine similarity loss degraded more than teacher initialization, showing that the former method had more impact on improving DistilHuBERT, while the latter was less effective.

\subsection{Layer Selection}
\begin{figure}[t]
    \centering
    \includegraphics[width=\linewidth]{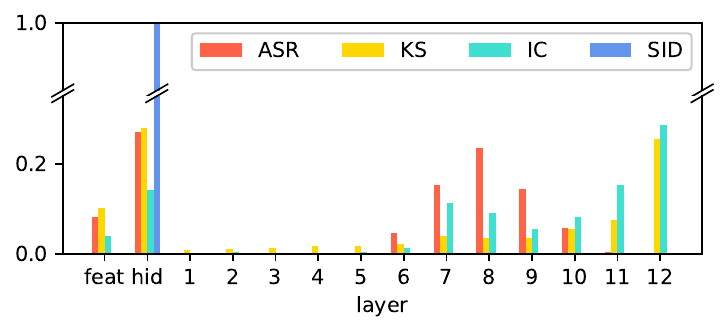}
    \vspace{-25pt}
    \caption{
        Weights of the feature summarization in several SUPERB tasks after training on DistilHuBERT representations (predicting all hidden layers in HuBERT).
        \texttt{feat} and \texttt{hid} respectively denotes the CNN's and transformer's outputs.
    }
    \label{fig:task_weights}
    \vspace*{-8pt}
\end{figure}

\begin{table}[t]
    \caption{
        Results of predicting different HuBERT's hidden layers.
    }
    \vspace{2pt}
    \label{tab:layer}
    \centering
    \begin{tabular}{lccc}
        \toprule
        & IC & SID & ASR \\
        \cmidrule{2-4}
        Predicted Layers & Acc$\uparrow$ & Acc$\uparrow$ & WER$\downarrow$ \\
        \midrule
        4, 8, 12 & 94.99 & 73.54 & 13.34 \\
        \midrule
        4 & 79.09 & 76.85 & 14.90 \\
        8 & 96.89 & 61.52 & 12.28 \\
        12 & 95.52 & 65.14 & 13.48 \\
        4, 8 & 91.67 & 75.36 & 13.73 \\
        4, 12 & 91.88 & 74.48 & 14.03 \\
        8, 12 & 96.34 & 65.30 & 13.03 \\
        \bottomrule
    \end{tabular}
    \vspace*{-10pt}
\end{table}

This section discussed the selection of HuBERT's hidden layers for DistilHuBERT to learn.
Because each HuBERT layer has different information, we first inspected each layer’s importance for
several downstream tasks.
We trained a DistilHuBERT model with 12 prediction heads to predict all HuBERT layers.
The outputs of DistilHuBERT's hidden layers and all prediction heads were weighted and summed by a set of learnable weights, where the weights were jointly trained with the downstream model.
The summarized outputs were used for all downstream tasks with separate weights.
The weights were normalized by the averaged $\ell2$ norm of each layer's output.
We visualized the learned weights in Fig. 3, where larger values indicate higher importance.
We only showed ASR, KS, IC, and SID because they represent recognition, detection, semantics, and speaker identity tasks, while others showed similar trends.

Fig.~\ref{fig:task_weights} showed features before the prediction heads (\texttt{hid}) were useful, especially for SID.
Results corroborated our hypothesis that the shared representation for multiple heads carried rich information.
Among the head outputs, the heads for the 6$^\text{th}$ to 12$^\text{th}$ layers were also crucial to ASR, KS, and IC because they had content and semantic information.
With the above findings, we made DistilHuBERT predicting the 8$^\text{th}$ and 12$^\text{th}$ HuBERT layers.
DistilHuBERT additionally predicted the 4$^\text{th}$ layer because we hypothesized that the bottom layers preserved speaker identity.

To justify our selection method, we conducted experiments by predicting different HuBERT layers as shown in Table \ref{tab:layer}.
First, for predicting only one HuBERT layer, the evaluation scores among the three tasks were imbalanced.
Predicting the 4$^\text{th}$ layer performed the worst in IC and ASR but better in SID than predicting three layers, corroborating our hypothesis that the bottom layers offered speaker identity.
Then, predicting two of the selected layers, the downstream tasks' scores were also imbalanced.
Therefore, the three selected layers offered a good combination for learning representations with less biased information.

\subsection{Knowledge Distillation with Different Datasets}
\label{subsec:diff_data}
\begin{table}[t]
    \caption{
        Pre-training DistilHuBERT with different datasets.
    }
    \vspace{2pt}
    \label{tab:data}
    \centering
    \begin{tabular}{lcccc}
        \toprule
        & Data Size & IC & SID & ASR \\
        \cmidrule{2-5}
        Dataset & hours & Acc$\uparrow$ & Acc$\uparrow$ & WER$\downarrow$ \\
        \midrule
        LibriSpeech & 960 & 94.99 & 73.54 & 13.34 \\
        \midrule
        LibriSpeech & 100 & 93.17 & 69.46 & 14.77 \\
        WSJ & 81 & 90.22 & 64.14 & 15.59 \\
        AISHELL-1 & 150 & 87.29 & 67.65 & 16.42 \\
        \bottomrule
    \end{tabular}
    \vspace*{-8pt}
\end{table}

In practice, the pre-training data for a self-supervised teacher model might be inaccessible or too large to reuse for distillation.
This section inspected the generalizability and flexibility of DistilHuBERT by training it with small or out-of-domain datasets.
We used the 100-hour subset of LibriSpeech, 81-hour WSJ, and the Mandarin AISHELL-1 corpora to train DistilHuBERT by the proposed knowledge distillation approach.
The results are shown in Table \ref{tab:data}.

First, training with the 100-hour LibriSpeech, DistilHuBERT degraded slightly but retained better performance than many other methods (in Table \ref{tab:superb}), indicating that using a smaller dataset to distill knowledge was sufficient.
Next, the smaller WSJ corpus showed more degradation; perhaps the WSJ's averaged utterance duration was shorter than LibriSpeech \cite{chang2021lifelong} or simply because it was small. 
Furthermore, training with the Mandarin AISHELL-1 corpus offered better SID than WSJ because speaker identity was independent of language and AISHELL had more data than WSJ.
In contrast, AISHELL performed the worst in IC and ASR because of the language mismatch.
Although the performance was affected by training data, DistilHuBERT offered better performance than many other SSL methods, even with smaller datasets for distillation.

\section{Conclusion}

This paper proposes DistilHuBERT, a novel framework to layer-wise distill knowledge from HuBERT.
DistilHuBERT retained most of HuBERT's performance and had only 25\% of its size.
We provided comprehensive analyses of the proposed methods and demonstrated the DistilHuBERT's flexibility and generalizability.
Our method can be easily applied to more powerful SSL models.

\bibliographystyle{IEEEtran}
\bibliography{refs}

\end{document}